\documentclass[sigconf]{acmart}
\usepackage{graphicx}
\usepackage{array} 
\usepackage{makecell}
\usepackage{ulem}
\usepackage{enumitem}
\usepackage{flushend}
\usepackage{multirow}
\AtBeginDocument{%
  }

\copyrightyear{2023}
\acmYear{2023}
\setcopyright{acmlicensed}\acmConference[MM '23]{Proceedings of the 31st
ACM International Conference on Multimedia}{October 29-November 3,
2023}{Ottawa, ON, Canada}
\acmBooktitle{Proceedings of the 31st ACM International Conference on
Multimedia (MM '23), October 29-November 3, 2023, Ottawa, ON, Canada}
\acmPrice{15.00}
\acmDOI{10.1145/3581783.3612472}
\acmISBN{979-8-4007-0108-5/23/10}

\begin{document}

\title{Towards Deconfounded Image-Text Matching with Causal Inference}
\author{Wenhui Li}
\email{liwenhui@tju.edu.cn}
\affiliation{%
\institution{Tianjin University}
\state{Tianjin}
\country{China}}
\affiliation{
  \institution{Institute of Artificial Intelligence, Hefei Comprehensive National Science Center}
  \state{Hefei}
  \country{China}
}

\author{Xinqi Su}
\authornote{Corresponding author}
\email{suxinqi@tju.edu.cn}
\affiliation{%
  \institution{Tianjin University}
  \state{Tianjin}
  \country{China}}

\author{Dan Song}
\email{dan.song@tju.edu.cn}
\affiliation{%
  \institution{Tianjin University}
  \state{Tianjin}
  \country{China}}

\author{Lanjun Wang}
\email{wanglanjun@tju.edu.cn}

\affiliation{%
  \institution{Tianjin University}
  \state{Tianjin}
  \country{China}}

\author{Kun Zhang}
\email{zhangkun32@meituan.com}
\affiliation{%
  \institution{Meituan}
  \state{Beijing}
  \country{China}}

\author{An-An Liu}
\authornotemark[1]
\email{anan0422@gmail.com}
\affiliation{%
  \institution{Tianjin University}
  \state{Tianjin}
  \country{China}}
\affiliation{
 \institution{Institute of Artificial Intelligence, Hefei Comprehensive National Science Center}
  \state{Hefei}
  \country{China}
}

\renewcommand{\shortauthors}{Wenhui Li, et al.}

\begin{abstract}
Prior image-text matching methods have shown remarkable performance on many benchmark datasets, but most of them overlook the bias in the dataset, which exists in intra-modal and inter-modal, and tend to learn the spurious correlations that extremely degrade the generalization ability of the model. Furthermore, these methods often incorporate biased external knowledge from large-scale datasets as prior knowledge into image-text matching model, which is inevitable to force model further learn biased associations. To address above limitations, this paper  firstly utilizes Structural Causal Models (SCMs) to illustrate how intra- and inter-modal confounders damage the image-text matching. Then, we employ backdoor adjustment to propose an innovative Deconfounded Causal Inference Network (DCIN) for image-text matching task. DCIN (1) decomposes the intra- and inter-modal confounders and incorporates them into the encoding stage of visual and textual features, effectively eliminating the spurious correlations during image-text matching, and (2) uses causal inference  to mitigate biases of external knowledge. Consequently, the model can learn causality instead of spurious correlations caused by dataset bias. Extensive experiments on two well-known benchmark datasets, \textit{i.e.}, Flickr30K and MSCOCO, demonstrate the superiority of our proposed method.
\end{abstract}

\ccsdesc[500]{Information systems~Multimedia and multimodal retrieval}
\keywords{Image-Text Matching, Causality, Backdoor Adjustment}



\maketitle

\section{Introduction}
Image-text matching is a crucial task that devotes to bridging the semantic gap between computer vision and natural language processing.
It aims to align the semantics of image-text pairs, and matches images with their semantically consistent texts, or matches texts with their corresponding images. Recent studies have achieved impressive performance on image-text matching by focusing on two key aspects: (i) precise measurement of semantic similarity between the features of image-text pairs, and (ii) introducing external knowledge to enhance the capabilities of  cross-modal representation and semantic matching.  

\begin{figure*}[htp]
  \includegraphics[width=0.91\textwidth]{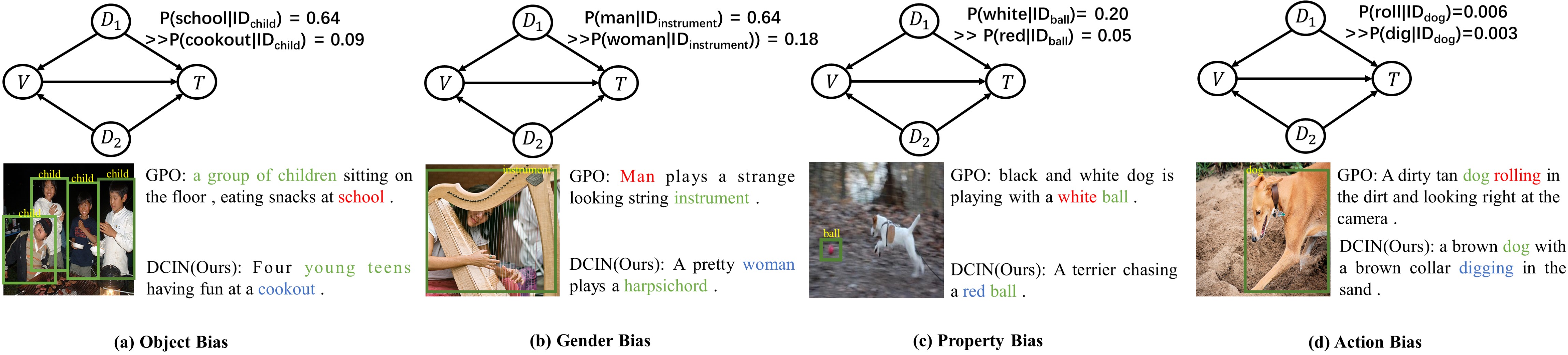}
  \caption{An illustration of spurious correlation in image-to-text retrieval. $D_1$, $D_2$ denote the visual and linguistic confounders, respectively. The probability represents the percentage of co-occurrence of two concepts in the training set, where ID denotes the visual concept. The top result is obtained by GPO, and the bottom is searched by our DCIN. Red and blue indicate wrong and correct words, respectively. Green boxes represent visual concepts that may have caused errors, with green words in the text corresponding to those visual concepts.}
  \label{fig:Motivation}
\end{figure*}
In the first aspect, the existing methods can be broadly categorized as global-level matching and local-level matching \cite{zhang2022negative}. The former projects holistic features of image and text into a common space to compute their semantic similarity \cite{chen2021learning,zeng2021softmax,li2019visual,yan2021discrete}, while the latter considers the semantic correspondence between specific visual and textual elements, such as region or word \cite{karpathy2014deep,lee2018stacked}. The attention-based local-level matching approach was proposed and quickly became the most popular framework for image-text matching by utilizing the attention mechanism to align the common semantics of image-text pairs \cite{lee2018stacked,zhang2022negative,pei2022scene,diao2021similarity,zhang2020context,qu2020context}, resulting in more comprehensive correspondences and advanced performance. Moreover, Chen et al. \cite{chen2021learning} discovered that selecting appropriate Generalized Pooling Operator (GPO) to adaptively pool multiple fragments~(regions or words) can lead to the more efficient and remarkable performance in global-level matching. In the second aspect, current methods aim to improve the ability to reason the relationships between image and text by utilizing co-occurrence probabilities of concepts from large-scale datasets as external knowledge for the model. For example, Wang et al. \cite{wang2020consensus} proposed a Consensus-aware Visual-Semantic Embedding approach that used the co-occurrence probability to infer higher-level relationships between image and text. Similarly, Feng et al. \cite{feng2023mkvse} also tended to establish reliable image-text connections at a higher level by utilizing the co-occurrence probability.

Despite remarkable performance achieved in image-text matching, current methods often overlook the intra- and inter-modal biases that exist in the dataset used for training and introducing external knowledge, which severely impair the model's generalization ability. For instance, if two concepts within the intra-modal frequently co-occur, the model may incorrectly assume once a concept appears, another concept will also  occur. Similarly, if regions and words in the inter-modal are not firmly related but frequently co-occur, the model may consider them overly related when calculating similarity. Fig.\ref{fig:Motivation}(a) illustrates an demonstration of object bias in image-retrieve-text task, where the GPO model \cite{chen2021learning} retrieved a caption "a group of children sitting on the floor, eating snacks at school" for a given image of "Some children having a good time at a cookout.", while the semantic "school" is not present in the image. This could be due to the frequent co-occurrence of "child" and "school" within intra- and inter-modal, which is far greater than the co-occurrence of "child" and "cookout". Likewise, gender bias (Fig.\ref{fig:Motivation}(b)), property bias (Fig.\ref{fig:Motivation}(c)), and action bias (Fig.\ref{fig:Motivation}(d)) also exist in image-text matching. 

In order to address the aforementioned limitations, we propose using causal inference to eliminate spurious correlations caused by dataset bias during model learning process. Specifically, we initially utilize structural causal models (SCMs) to demonstrate how intra- and inter-modal biases of training set damage the performance of image-text matching. Subsequently, we employ backdoor adjustment to propose an innovative Deconfounded Causal Inference Network (DCIN) for image-text matching, which mainly consists of two steps: (i) we decompose the intra- and inter-modal confounders of the training set in the image-to-text (text-to-image) task into visual confounder (linguistic confounder) and linguistic confounder (visual confounder), and utilize them to implement  causal intervention in the encoding of image and text features, efficiently eliminating the spurious correlations stemming from training set and forcing the model to learn causal knowledge instead of common co-occurrence. (ii)  we use causal probability estimation to mitigate bias when introducing external knowledge. As shown in Figure.\ref{fig:Motivation}(a), DCIN can accurately match the text " Four young teens having fun at a cookout". The main contributions of our work are summarized as follows:
\begin{itemize}
\item We introduce the causal inference into image-text matching and construct structural causal model to eliminate the intra-modal and inter-modal spurious correlation of semantics caused by dataset bias.
\item We propose an innovative deconfounded causal inference network to incorporate the confounders into  visual and textual feature learning, and utilize the backdoor adjustment to address the co-occurrence interference in the database as well as  external knowledge.
\item Extensive qualitative and quantitative experiments on two benchmarks, \textit{i.e.}, Flickr30K and MSCOCO, show that the superiority of our proposed method. 
\end{itemize}

\section{Related Work}

\subsection{Image-text Matching}
Image-text matching methods can be broadly classified into two major categories: global-level matching and local-level matching. 

The former measures overall semantic alignment, while the latter focuses on fine-grained alignment between local segments of image-text pairs. Karpathy et al. \cite{karpathy2015deep} were the first to attempt fine-grained semantic reasoning for images and texts. Subsequently, Lee et al. \cite{lee2018stacked} employed the Stacked Cross Attention Network (SCAN) to establish correspondences between regions and words, enabling each region to attend to multiple words and vice versa. SCAN's success led to attention-based local-level matching methods becoming the predominant paradigm for image-text matching, with various variants developed \cite{zhang2022negative,diao2021similarity,zhang2020context}. In these methods, a crucial research direction is to design effective cross-modal alignment mechanisms, such as object-relational alignment \cite{liu2020graph,zhang2020context,chen2023more}, gating mechanism fusion \cite{chen2021adapted,qu2020context,wang2019camp}, and elimination of irrelevant fragments \cite{zhang2022negative,diao2021similarity,liu2019focus}, etc. Global-level matching methods typically project image-text pairs to an embedding space to measure semantic similarity \cite{zeng2021softmax,chen2021learning,faghri2017vse++}. Early optimization used ranking losses for improved matching. For example, the hinge-based triplet loss was employed to ensure that the similarity scores of corresponding pairs are higher than non-corresponding ones \cite{klein2015associating}. Faghri et al. \cite{faghri2017vse++} optimized the triplet loss by focusing on the hardest mismatched pairs during training. Lately, global-level matching approaches have focused on aggregation strategies for visual and textual representations. Chen et al. \cite{chen2021learning} proposed an adaptive global pooling function to learn the best pooling strategy between features. Zeng et al. \cite{zeng2021softmax} introduced the softmax pooling strategy to enhance visual semantic embedding, resulting in more efficient and advanced performance.

Recently, researchers utilize internal and external knowledge to enhance associations between image and text pairs \cite{shi2019knowledge,wang2020consensus,huang2022mack}. For example, Wang et al. \cite{wang2020consensus} utilized internal knowledge, while Feng \cite{feng2023mkvse} and Shi \cite{shi2019knowledge} et al.  incorporated external knowledge to enhance cross-modal representation and semantic matching, which obtained better alignment.
\subsection{Causal Inference}
In recent years, researchers have integrated causal inference into computer vision and natural language processing \cite{chalupka2017causal,lopez2017discovering,zhang2020causal}, enabling DNNs to learn causal effects and significantly improving performance in some areas such as image classification \cite{lopez2017discovering,zhang2021deep}, visual question answering \cite{pan2022causal,pan2021distilling}, and image captioning \cite{yang2021deconfounded,liu2022show}.

Causal inference methods can be categorized into front-door  and backdoor adjustment \cite{glymour2016causal,pearl2018does,yang2021deconfounded}. Backdoor adjustment stratifies confounders into different levels to deconfound. Wang et al. \cite{wang2020visual} proposed Visual Commonsense Region-based Convolutional Neural Network, which employed backdoor adjustment to improve visual feature representation learning. Zhang et al. \cite{zhang2020devlbert} utilized backdoor adjustment to mitigate high likelihood of co-occurrence between visual and textual tokens, showing that reducing dataset bias can enhance generalization. Liu et al. \cite{liu2022show} used backdoor adjustment to disentangle region visual features and deconfound visual and linguistic pseudo-correlations in image captioning. Front-door adjustment aims to construct an additional mediator between cause and effect relationship to transmit knowledge. Yang et al. \cite{yang2021deconfounded} used the front-door adjustment to deconfound in image captioning. Yang et al. \cite{yang2021causal} improved the attention mechanism with the front-door adjustment to remove spurious correlations.

In image-text matching task, confounders exist in both inter- and intra-modal aspects. Due to bi-directional retrieval, the features of image-text pairs can be easily interfered by these confounders, thus deconfounding is more challenging than other tasks.

\begin{figure}[htp]
\includegraphics[width=0.95\linewidth]{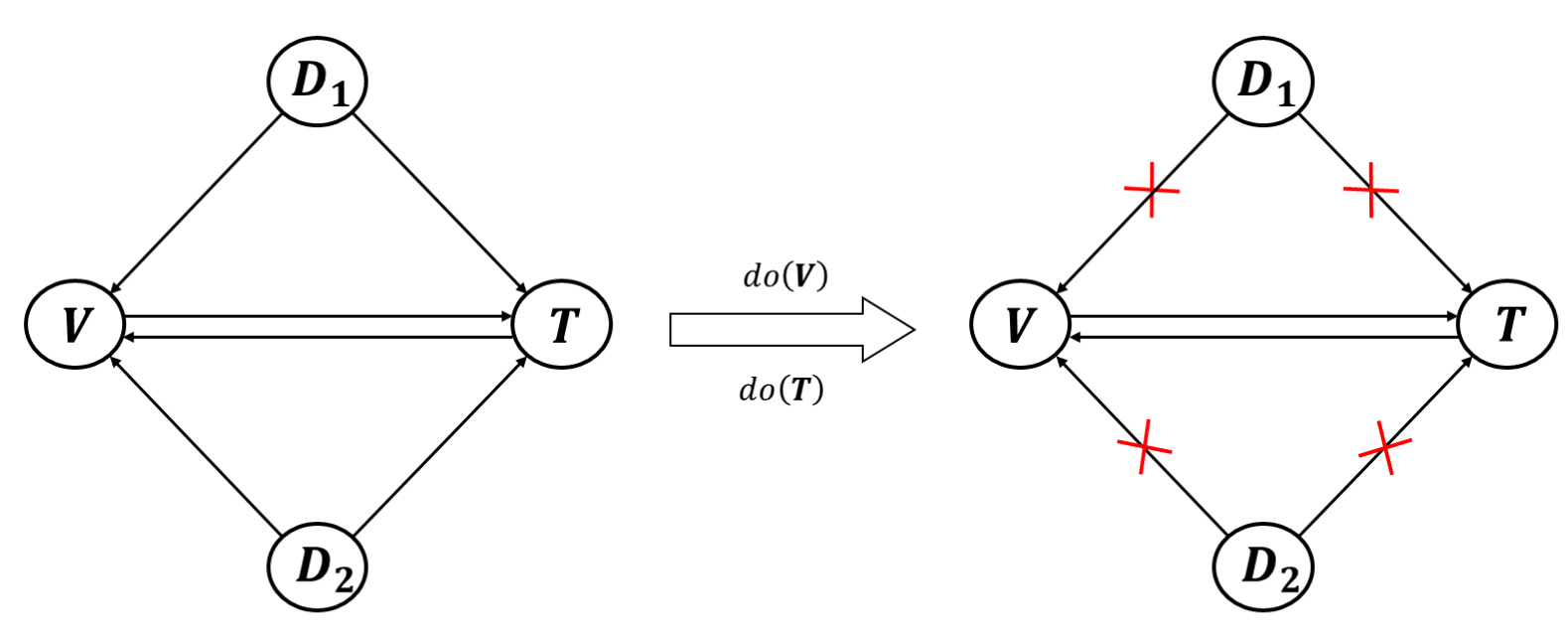}
\caption{Left: structural causal model (SCM) for image-text matching. Right: intervention with backdoor adjustment. The direction of an edge in the SCM indicate solely the causal relationship, directing from the cause to the effect. The variables $V$, $T$, $D_1$, $D_2$ refer to image, text, visual confounder and linguistic confounder, respectively.}
\label{fig:scm}
\end{figure}

\section{METHODOLOGY} 

\begin{figure*}[htp]
  \includegraphics[width=0.95\textwidth]{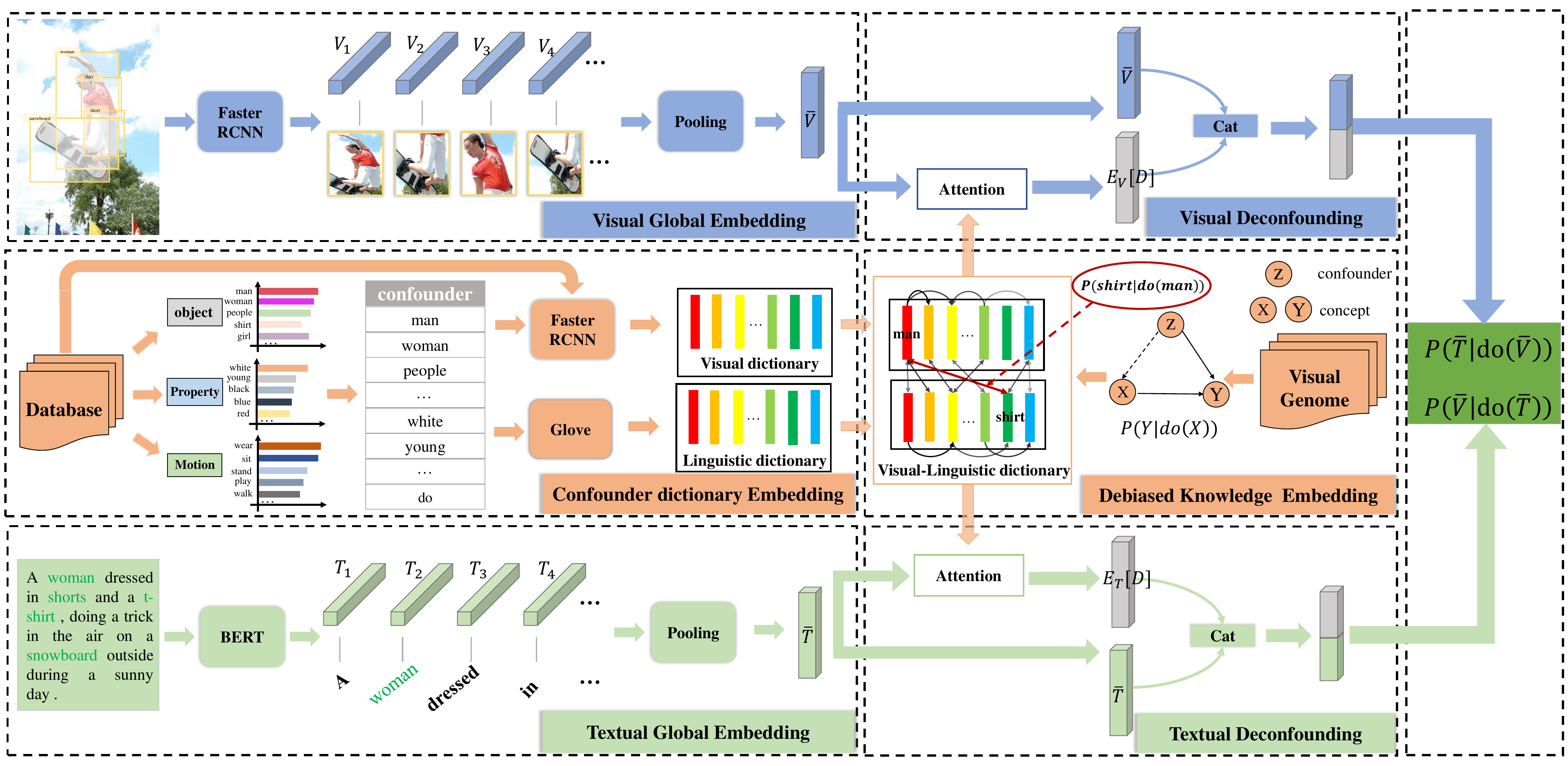}
  \caption{Illustration of the Deconfounded Causal Inference Network, which consists of three main modules designed to eliminate the spurious correlations between image and text features. (1) The Confounder Dictionary Embedding module constructs approximate visual and linguistic confounder dictionary.(2) The Debias Knowledge Embedding module incorporates debiased knowledge from the Visual Genome dataset to enhance the model's matching ability. (3) The Visual-Textual Feature Deconfound module uses causal inference to confront visual and linguistic confounders and eliminates the spurious correlations during image-text matching to obtain accurate image or text matching results. }
  \label{fig:frame}
\end{figure*}
In this section, we first construct a structure causal model for image-text matching and explain how confounder can lead to spurious correlations in Sec. \ref{M1}, and then describe how to use the backdoor adjustment to eliminate the spurious correlations in Sec. \ref{M2}. Finally, we provide a detailed description of our method and how it generates unbiased solution in Sec. \ref{M3}.

\subsection{SCM for Image-text matching}\label{M1}
Structure causal model(SCM) is formulated to analyze the causal relationship where a variable is defined as a confounder when it exerts an influence on two or more other variables simultaneously \cite{yang2021deconfounded,chen2022towards}. To simplify, we denote $V$ as the input image or image feature and $T$ as the input text or text feature. In Fig. \ref{fig:scm}, we present a SCM constructed for image-text matching, depicting the relationships among image feature $V$, text feature $T$, visual confounder $D_1$, and linguistic confounder $D_2$, where the direction of an edge indicates causality, e.g., $D_1 \rightarrow V$ means that $D_1$ causes $V$. Next, we will introduce the structure causal model in detail. 

\textbf{$V \rightarrow T$, $T \rightarrow V$:} The causal effect $V \rightarrow T$ indicates that the visual features contribute to matching their corresponding texts, and $T \rightarrow V$ indicates that the text features contribute to matching their corresponding images.

\textbf{$D_1 \rightarrow V$, $D_1 \rightarrow T$, $D_2 \rightarrow V$, $D_2 \rightarrow T$:} The confounder can negatively affect the attended features. We denote the causal effect of $D_1$ on $V$ as $D_1 \rightarrow V$ because the frequently appearing visual contexts can severely affect the attended visual features during training \textbf{(intra-modal)}. $D_1 \rightarrow T$ means that the visual contexts can affect the frequency of textual words~\textbf{(inter-modal)} \cite{liu2022show}. Similarly, $D_2 \rightarrow V$ implies that the frequently appearing linguistic contexts can directly affect the frequency of visual regions~\textbf{(inter-modal)}, and $D_2 \rightarrow T$ indicates that the linguistic contexts can strongly impact the attended textual features by co-occurring \textbf{(intra-modal)}.

\textbf{$V \leftarrow D_1 \rightarrow T$, $V \leftarrow D_2 \rightarrow T$, $T \leftarrow D_1 \rightarrow V$, $T \leftarrow D_2 \rightarrow V$ \textbf{(backdoor path)} :} The backdoor path between $V$ and $T$ is defined as any path from $V$ to $T$ with an arrow pointing towards $V$. The existence of confounders $D_1$ and $D_2$ create multiple backdoor paths between $V$ and $T$,  which result in taking the observational likelihood $P(T\mid V)$ or $P(V \mid T)$ as the training target and generating spurious correlations between $V$ and $T$, decreasing the network's generalization.

To illustrate how backdoor paths in image-text matching can lead to the spurious correlations between image and text features, we use the Bayes' theorem to decompose $P(T \mid V)$ as an example:
\begin{equation} \label{eqn1}
\begin{aligned}
P(T \mid V)= & \sum_{d_2} P\left(d_2 \mid V\right) \cdot \\ & \sum_{d 1} P\left(T \mid V, d_1, d_2\right) P\left(d_1 \mid V\right).
\end{aligned}
\end{equation}
The confounders $D_1$ and $D_2$ can introduce bias through their respective conditional probabilities $P(d_1 \mid V)$ and $P(d_2 \mid V)$, where $d_1$ and $d_2$ represent specific factors of $D_1$ and $D_2$. For example, if there is a dataset bias that leads to $P(d_{ID_{school}} | ID_{child}) \approx 1$ or $P(d_{school} | ID_{child}) \approx 1$,  then $P(T \mid ID_{child})$ will become $P(T \mid ID_{child}, d_{ID_{school}})$ or $P(T,d_{school} \mid ID_{child})$, resulting in a spurious correlation towards the region or word "\textit{school}".

\subsection{Intervention with Backdoor Adjustment}\label{M2}
We propose using backdoor adjustment in causal inference to eliminate dataset bias. We employ backdoor adjustment to block the backdoor path, as shown in Fig.\ref{fig:scm} (right). The model's implicit training target is set as $P(T \mid do(V))$ and $P(V \mid do(T))$, instead of the regular target for image-text matching. Specifically, we  intervene on variable $V$ and utilize backdoor adjustment to achieve $P(T \mid do(V))$ as follows:
\begin{equation} \label{eqn2}
\begin{aligned}
P(T \mid d o(V))= & \sum_{d_2} P\left(d_2\right) \cdot \\ & \sum_{d 1} P\left(T \mid V, d_1, d_2\right) P\left(d_1\right) ,
\end{aligned}
\end{equation}
where $do( \cdot )$ is the \textit{do-operator}, denoting the experimental intervention. Using the intervention probability in Eq.\eqref{eqn2}, the model is compelled to understand the impact of each confounder and learn the causal effect of $V \rightarrow T$, rather than the spurious correlations generated by visual confounder $D_1$ and linguistic confounder $D_2$.

\subsection{DCIN}\label{M3}
Our proposed DCIN is depicted in Fig.\ref{fig:frame}. First, we extract image and text global features in the "Visual and Textual Global Embedding". Then, we create visual and linguistic  confounder dictionaries and incorporate debiased external knowledge based on causal probability estimation $P(Y \mid do(X))$ in the "Confounder dictionary Embedding" and "Debiased Knowledge  Embedding" modules. Finally, we eliminate the spurious correlations between visual and textual features in the "Visual Deconfound" and "Textual Deconfound" modules before computing these matching probability.
\subsubsection{ Visual and Textual Global Embedding.} 
For an input image, we follow the approaches of \cite{zhang2022negative,chen2021learning} to extract salient regions using Faster R-CNN \cite{ren2015faster}, which pre-trained on the Visual Genome \cite{krishna2017visual}, by selecting the top-$n$ ($n = 36$) regions with the highest confidences. Then, we encode region-level image features $F=\left\{f_1, \ldots, f_i, \ldots, f_n\right\}$ from these regions using ResNet101 \cite{he2016deep} pre-trained on ImageNet \cite{deng2009imagenet}. Finally, we employ a fully connected layer to map these features to a $D$-dimensional space to obtain image features $V=\left\{v_1, \ldots, v_i, \ldots,v_n\right\}$. In order to acquire  global embedding $\overline V$, we apply generalized pooling operator (GPO) \cite{chen2021learning} on  features $V$.  For an input text, we utilize pre-trained Bert \cite{devlin2018bert} as text encoder to extract the word representation $C=\left\{c_1, \ldots, c_i, \ldots, c_m\right\}$. Subsequently, we project 
 the $C$ into a $D$-dimensional space using a fully connected layer to obtain the text features $T=\left\{t_1, \ldots, t_i, \ldots, t_m\right\}$.  Similarly, GPO is also utilized to acquire  global embedding $\overline T$ for text feature $T$.
 
\subsubsection{Confounder dictionary Embedding.} 
To construct the necessary confounders for backdoor adjustment, we create approximate visual confounder dictionary $D_1$ and linguistic confounder dictionary $D_2$. Due to the impracticality of focusing on all concepts in the training set and identifying all confounders, we extract the top-$k$ semantic concepts in text and visual domains and remove non-informative words such as \textit{"is"} and \textit{"a"}. We select the dictionary containing \textit{object, gender, property, action} concepts to address their corresponding biases.

To construct the linguistic dictionary, we use pre-trained Glove \cite{pennington2014glove} word embedding to represent the selected text semantic concepts and create the linguistic matrix ${\mathbf{g}^t} \in \mathbb{R}^{k \times d_t}$. As for the visual dictionary, we adopt the approach from previous work \cite{feng2023mkvse} and represent each visual concept as a region-level average ROI feature of the same image, constructing the visual matrix ${\mathbf{g}^v} \in \mathbb{R}^{k \times d_v}$.
\begin{equation} \label{eqn5}
\mathbf{g}_i^v=\frac{1}{h_i} \sum_{j=1}^{h_i} \frac{1}{R} \sum_{r=1}^R \mathbf{v}_{j, r}   ,
\end{equation}
where $h_i$ denotes the number of images containing the $i-th$ concept, and $\mathbf{v}_{j, r}$ denotes the feature of the $r-th$ region of the $j-th$ image. Next, we utilize two linear projections, ${W^v} \in {\mathbb{R}^{{d_v} \times d}}$ and ${W^g} \in \mathbb{R}{^{{d_t} \times d}}$, to map ${V^r}$ and ${g^r}$ to $D_1$ and $D_2$, respectively. \textit{i.e.}, ${D_1} = {W^v}{\mathbf{g}^v}$, ${D_2} = {W^g}{\mathbf{g}^t}$.
 \begin{figure}[htp]
  \includegraphics[width=0.95\linewidth]{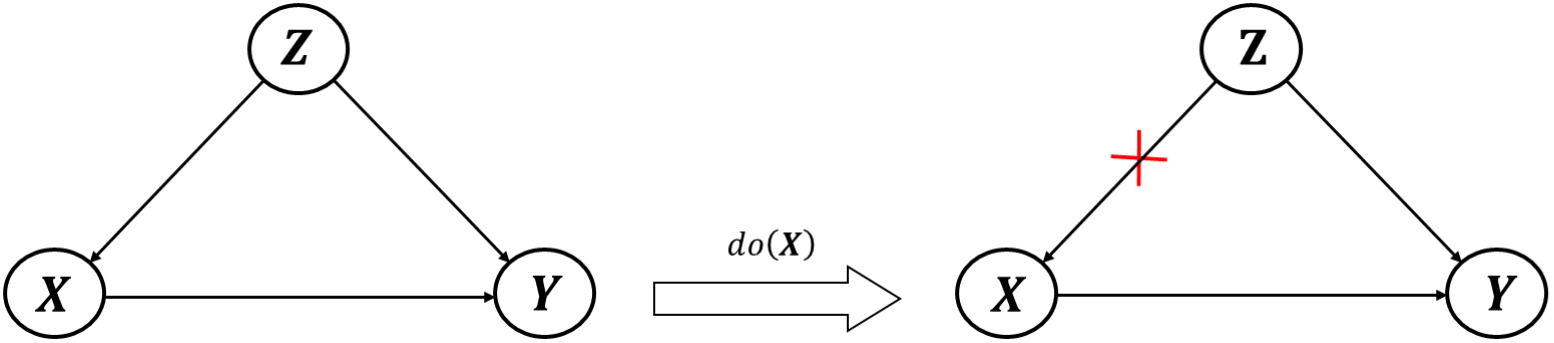}
  \caption{Left: structural causal model (SCM) for incorporating external knowledge. Right: intervention with backdoor adjustment. The variables $X$, $Y$ indicate semantic concepts, and $Z$ denotes confounder.}
  \label{fig:scm2}
\end{figure}
\subsubsection{Debias Knowledge Embedding.} 
Existing methods often leverage external knowledge from large-scale datasets to enhance the reasoning ability of image-text matching models at higher levels. They tend to capture the relationships between concepts $X$ and $Y$ using their co-occurrence probabilities $P(Y\mid X)$. However, these approaches often introduce dataset bias, leading to spurious correlations between different concepts \cite{glymour2016causal}. We conduct a toy experiment to illustrate the impact of bias on the relationship between semantic concepts. Surprisingly, we find that given the concept "\textit{plate}," the probability of "\textit{fork}" being present is high, \textit{i.e.}, $P(fork \mid plate)= 8.94\%$, even though it does not have a strong relationship with the concept "\textit{fork}". Fig.\ref{fig:scm2} (left) illustrates a SCM used to introduce external knowledge, where $P(Y \mid X)$ can be represented as:
\begin{equation} \label{eqn6}
P(Y \mid X)=\sum_z P(Y \mid X, z) P(z \mid X)=\frac{P(Y, X)}{P(X)},
\end{equation}
where $Z$ denotes the confounder. When $Z$ is observed, we usually observe both $X$ $(Z \rightarrow X)$ and $Y$ $(Z \rightarrow Y)$. However, when only $X$ is observed, $Y$ rarely appears, indicating that $Z$ is the confounders between $X$ and $Y$ $(X \leftarrow Z \rightarrow Y)$. To address this problem, we use the backdoor adjustment to control the confounder effect of $Z$ and mitigate the spurious correlation between $X$ and $Y$. The adjusted formula is as follows:
\begin{equation} \label{eqn7}
P(Y \mid d o(X))=\sum_z P(Y \mid X, z) P(z)=\sum_z \frac{P(Y, X, z) P(z)}{P(X, z)},
\end{equation} 
where $Z$ is the visual (linguistic) confounder if $X$, $Y$ are visual (textual) concepts, and $Z$ is the visual and linguistic confounders if $X$, $Y$ are concepts of different modalities. To prevent data leakage in the test set, we select image-text pairs from the Visual Genome dataset \cite{krishna2017visual} that are also present in either the MSCOCO \cite{lin2014microsoft} or Flickr 30K \cite{plummer2015flickr30k} training sets as external knowledge, we utilize multiple stacked graph convolutional networks (GCNs) to model the interdependencies among the concepts in the confounder dictionary. Firstly, we construct the concept relationship matrix $\mathbf{E}$, where $\mathbf{E}_{i j}=P\left(Y_j \mid d o\left(X_i\right)\right)$. Then, we concatenate the visual dictionary $D_1$ and linguistic dictionary $D_2$, and reason the relationship between them. The process is illustrated below:
\begin{equation} \label{eqn8}
\begin{aligned} & \mathbf{H}^{(0)}=[{D_1, D_2}] \\ & \mathbf{H}^{(l+1)}=\rho\left(\widetilde{\mathbf{A}} \mathbf{H}^{(l)} \mathbf{W}^{(l)}\right) + \mathbf{H}^{(0)}, \end{aligned}
\end{equation}
where $\tilde{\mathbf{A}}=\mathbf{D}^{-\frac{1}{2}} \mathbf{E} \mathbf{D}^{-\frac{1}{2}}$ denotes normalized symmetric matrix, ${W}^{(l)}$ is learnable weight matrix, and $\rho$ is LeakyReLU activation function. We can obtain the final confounder dictionary $D$ from the output of the last layer of the GCN.

\subsubsection{Visual-Textual Feature Deconfound.} 
Our goal is to transform the hidden probabilities $P( T \mid  V)$ and $P( V \mid  T)$ into $P( T \mid do( V))$ and $P( V \mid do( T))$ during the model learning process. We can compute Eq.\eqref{eqn2} as follows:
\begin{equation} \label{eqn9}
P({T} \mid d o({V}))=\mathbb{E}_{D_1} \mathbb{E}_{D_2}\left[P\left(\overline{T} \mid \overline{V}, d_1, d_2\right)\right].
\end{equation}
It can be observed that it is the expected value of $P\left(\overline{T} \mid \overline{V}, d_1, d_2\right)$ according to the $D_1$ and $D_2$, and following the previous works \cite{wang2020consensus,yang2021causal}, we can represent $P(T\mid do( V))$ as a network, where $P\left(\overline{T} \mid \overline{V}, d_1, d_2\right)$ can be implemented as:
\begin{equation} \label{eqn9_1}
P\left(\overline T \mid \overline V, d_1, d_2\right)=g\left(\overline {V}, \overline {T}, d_1, d_2\right),
\end{equation}
where  $g(\cdot)$ is a linear model to compute the match probability of the image-text pair, e.g., $g\left(\overline{V}, \overline{T}, d_1, d_2\right)= \left(Q_1 \overline{V}+Q_2 d_1  +Q_3 d_2\right) \times \overline{T}$, where $Q_1,Q_2,Q_3 \in \mathbb{R}^{d \times d} $ denote the learnable weights. However, this approach also presents a challenge that in order to compute $P( T \mid do(V))$, we need to sample a large number of outputs from the network. To address this issue, we absorb the expectations into the linear model $g(\cdot)$ because the linear projection of the expectation of a variable is equal to the expectation of the linear projection of that variable \cite{yang2021deconfounded}. Thus, the final equation can be expressed as:
\begin{equation} \label{eqn9_2}
\begin{aligned} P( {T} \mid \operatorname{do}( {V}))  & =\mathbb{E}_{D_1} \mathbb{E}_{D_2}\left[g\left(\overline {V},\overline {T}, d_1, d_2\right)\right] \\ & =g\left(\overline{V}, \overline{T},\mathbb{E}_{D_1}\left[D_1\right], \mathbb{E}_{D_2}\left[D_2\right]\right).\end{aligned}
\end{equation}

Likewise to the previous methods proposed in \cite{liu2022show,yang2021deconfounded}, we set $D_1$ and $D_2$ to be conditioned on the feature $\overline V$ to increase the representation power of the model as follows:
\begin{equation} \label{eqn9_3}
\begin{aligned} 
P({T} \mid d o(V))=g\left(\overline{V},\overline{T}, \mathbb{E}_{\left[D_1 \mid \overline V\right]}\left[D_1\right], \mathbb{E}_{\left[D_2 \mid \overline V\right]}\left[D_2\right]\right).
\end{aligned}
\end{equation}

To reduce the cost of sampling, we simultaneously sample $D_1$ and $D_2$ instead of sampling them separately, \textit{i.e.}, $D=[D_1, D_2]$, $\mathbb{E}_D^V[D] \approx  MLP(\operatorname{softmax}(D\overline{V}) D)$, $MLP(.)$ denotes the two-layer multi-layer perceptron. Note that $D$ can denote either the original confounder dictionary or an external knowledge-enhanced dictionary. Moreover, in image-text matching, there are some unobservable confounders and beyond visual and linguistic confounders, e.g., color, attributes, and nuanced scene contexts associated by them \cite{wang2020visual}. It is not reasonable to use the approximate visual and linguistic  confounder dictionaries alone. Therefore, we introduce a confidence level $\lambda$ for the  
approximate confounder dictionary. Empirically, we find that a simple weighting operation can achieve satisfactory results, and finally, $P(T \mid do(V))$ can be implemented as follows:
\begin{equation} \label{eqn10}
P(T \mid d o(V)) = g\left(\sqrt{(1-\lambda)} \overline{V}, \overline{T}, \sqrt{\lambda} \mathbb{E}_D^V[D]\right) .
\end{equation}

Similarly, for deconfound of text to retrieve image, we can implement it as follows:
\begin{equation} \label{eqn11}
P(V \mid d o(T)) = g\left(\sqrt{(1-\lambda)} \overline{T},  \overline{V},\sqrt{\lambda} \mathbb{E}_D^T[D]\right) .
\end{equation}

In order to achieve joint training of image and text matching, we directly compute their matching probability $S(V,T)$ between the deconfounded image and text features:
\begin{equation} \label{eqn12}
S=\sigma\left[\psi\left(\sqrt{(1-\lambda)} \overline{V}, \sqrt{\lambda} \mathrm{\mathbb{E}}_D^V[D]\right) \odot \psi\left(\sqrt{(1-\lambda)} \overline{T}, \sqrt{\lambda} \mathrm{\mathbb{E}}_D^T[D]\right)\right]  ,
\end{equation}
where $\odot$ is inner product, $\sigma$ denotes tanh activation function and $\psi$ is an embedding layer. 

\subsubsection{Objective Function.}\label{M4} 
To optimize the whole model, we utilize a hinge-based bidirectional triple loss \cite{faghri2017vse++} to enforce the matching probability of aligned image-text pairs to be a certain magnitude higher than that of unaligned pairs. The loss function is formulated as follows:
\begin{equation} \label{eqn_10}
L=[\alpha-S(V, T)+S(V, \hat{T})]_{+}+[\alpha-S(V, T)+S(\hat{V}, T)]_{+}  ,
\end{equation}
where $\alpha$ represents the margin factor,  ${[x]_ + } = max(x,0)$. $\hat{V}$ and $\hat{T}$  represent the hardest negatives samples in a mini-batch.

\begin{table}[]
\caption{Qualitative evaluation of bi-directional retrieval results on the Flickr30K test set. The best results are highlighted in bold, and the symbol * represents the ensemble model.}
\label{tab:1}
\setlength{\tabcolsep}{0.5mm}{
\begin{tabular}{lccccccc}
\toprule
\multicolumn{1}{c|}{} &
  \multicolumn{3}{c|}{Image-to-Text} &
  \multicolumn{3}{c|}{Text-to-Image} &
   \\ \cline{2-7}
\multicolumn{1}{c|}{\multirow{-2}{*}{Method}} &
  {\color[HTML]{000000} R@1} &
  {\color[HTML]{000000} R@5} &
  \multicolumn{1}{c|}{{\color[HTML]{000000} R@10}} &
  {\color[HTML]{000000} R@1} &
  {\color[HTML]{000000} R@5} &
  \multicolumn{1}{c|}{{\color[HTML]{000000} R@10}} & \multirow{-2}{*}{rSum} \\ \hline
\multicolumn{1}{l|}{SCAN*
\tiny
{(18)\cite{lee2018stacked}}} &
  67.4 &
  90.3 &
  \multicolumn{1}{c|}{95.8} &
  48.6 &
  77.7 &
  \multicolumn{1}{c|}{85.2} &
  465.0 \\
\multicolumn{1}{l|}{BFAN\tiny{(19)\cite{liu2019focus}}} &
  68.1 &
  91.4 &
  \multicolumn{1}{c|}{-} &
  50.8 &
  78.4 &
  \multicolumn{1}{c|}{-} &
  288.7 \\
\multicolumn{1}{l|}{CVSE\tiny{(20)\cite{wang2020consensus}}} &
  73.5 &
  92.1 &
  \multicolumn{1}{c|}{95.8} &
  52.9 &
  80.4 &
  \multicolumn{1}{c|}{87.8} &
  482.5 \\
\multicolumn{1}{l|}{IMRAM*\tiny{(20)\cite{chen2020imram}}} &
  74.1 &
  93.0 &
  \multicolumn{1}{c|}{96.6} &
  53.9 &
  79.4 &
  \multicolumn{1}{c|}{87.2} &
  484.2 \\
\multicolumn{1}{l|}{SGRAF*\tiny{(21)\cite{diao2021similarity}}} &
  77.8 &
  94.1 &
  \multicolumn{1}{c|}{97.4} &
  58.5 &
  83.0 &
  \multicolumn{1}{c|}{88.8} &
  499.6 \\
\multicolumn{1}{l|}{GPO\tiny{(21)\cite{chen2021learning}}} &
  81.7 &
  95.4 &
  \multicolumn{1}{c|}{97.6} &
  61.4 &
  85.9 &
  \multicolumn{1}{c|}{91.5} &
  513.5 \\
  \multicolumn{1}{l|}{\begin{tabular}[c]{@{}l@{}}GPO+\\ SoftPool\tiny{(21)\cite{zeng2021softmax}}\end{tabular}} &
  82.2 &
  96.0 &
  \multicolumn{1}{c|}{98.0} &
  62.6 &
  { 86.8} &
  \multicolumn{1}{c|}{{ 92.3}} &
  { 517.9} \\
\multicolumn{1}{l|}{CMCAN*\tiny{(22)\cite{zhang2022show}}} &
  79.5 &
  95.6 &
  \multicolumn{1}{c|}{97.6} &
  60.9 &
  84.3 &
  \multicolumn{1}{c|}{89.9} &
  507.8 \\
\multicolumn{1}{l|}{VSRN++\tiny{(22)\cite{li2022image}}} &
  79.2 &
  94.6 &
  \multicolumn{1}{c|}{97.5} &
  60.6 &
  85.6 &
  \multicolumn{1}{c|}{91.4} &
  508.9 \\
\multicolumn{1}{l|}{NAAF*\tiny{(22)\cite{zhang2022negative}}} &
  81.9 &
  { 96.1} &
  \multicolumn{1}{c|}{{ 98.3}} &
  61.0 &
  85.3 &
  \multicolumn{1}{c|}{90.6} &
  513.2 \\
\multicolumn{1}{l|}{GPO+CFM\tiny{(22)\cite{wei2022synthesizing}}} &
  { \underline{82.5}} &
  95.7 &
  \multicolumn{1}{c|}{98.1} &
  { 62.9} &
  86.2 &
  \multicolumn{1}{c|}{91.8} &
  517.2 \\
  \multicolumn{1}{l|}{MV-VSE\tiny{(22)\cite{li2022multi}}} &
  { 82.1} &
  95.8 &
  \multicolumn{1}{c|}{97.9} &
  { 63.1} &
  86.7 &
  \multicolumn{1}{c|}{92.3} &
  517.9 \\
  \multicolumn{1}{l|}{CMSEI*\tiny{(23)\cite{ge2023cross}}} &
  { 82.3} &
  \underline{96.4} &
  \multicolumn{1}{c|}{\underline{98.6}} &
  { \underline{64.1}} &
  \underline{87.3} &
  \multicolumn{1}{c|}{\underline{92.6}} &
  \underline{521.3} \\
 \hline
\multicolumn{1}{l|}{\textbf{DCIN(ours)}} &
  82.2 &
  96.0 &
  \multicolumn{1}{c|}{98.0} &
  62.9 &
  87.5 &
  \multicolumn{1}{c|}{92.5} &
  519.1 \\
\multicolumn{1}{l|}{\textbf{DCIN-ek(ours)}} &
  83.0 &
  96.4 &
  \multicolumn{1}{c|}{98.6} &
  63.3 &
  87.8 &
  \multicolumn{1}{c|}{92.4} &
  521.5 \\
\multicolumn{1}{l|}{\textbf{DCIN*(ours)}} &
  \textbf{83.6} &
  \textbf{97.0} &
  \multicolumn{1}{c|}{\textbf{98.6}} &
  \textbf{65.1} &
  \textbf{88.9} &
  \multicolumn{1}{c|}{\textbf{93.3}} &
  \textbf{526.5} \\ 
\bottomrule
\end{tabular}
}
\end{table}

\section{Experiments}
\subsection{Datasets and Implementation Details.} 
\textbf{Datasets. }
To evaluate the effectiveness of our proposed method, we conducted extensive experiments on two commonly  benchmark datasets. Flickr30K \cite{plummer2015flickr30k} comprises 310,000 images and 155,000 sentences, and we follow the uniform data protocol as previous works \cite{zhang2022negative}, where 1,000 images are used for testing, 1,000 for validation, and 29,000 for training. MSCOCO \cite{lin2014microsoft}  is composed of 123,287 images and 616,435 sentences, which are divided into 5,000 test images, 5,000 validation images, and 113,287 training images \cite{feng2023mkvse}. For MSCOCO, we conduct both average 5-fold cross-validation tests~(1K) and full image tests on the 5K test set.

\textbf{Evaluation Metrics. }We use $Recall@k$($R@k$, $K=1$,$5$,$10$) as the evaluation metric for image-text matching, $R@k$ represents the percentage of ground-truth in the top-$k$ matched results. To evaluate the overall performance, we calculate the sum of all $R@k$ as $rSum$.

\textbf{Implementation Details. }
We conducted all experiments using PyTorch V1.7.1 on an NVIDIA RTX 3090Ti GPU. The AdamW optimizer is used for model optimization with an initial learning rate of 0.0005 and a 10\% decay every 10 epochs. We apply a one-tenth learning rate for BERT fine-tuning. The model is trained on both datasets for 25 epochs with a mini-batch size of 128. The size of the visual and linguistic confounder dictionaries is set to 300 for each, with a ratio of 7:2:1 for object, property, and action concepts. The number of GCN layers is set to 1, and the margin hyperparameter $\alpha$ is selected as 0.2. The confidence level $\lambda$ for causality is set to 0.05, as discussed in Section \ref{Ablation Study}.
\begin{table}[]
\caption{Quantitative evaluation of bi-directional retrieval results on MS-COCO 1K and 5K test set. The best results are highlighted in bold, and the symbol * represents the ensemble models.}
\label{tab:2}
\setlength{\tabcolsep}{0.3mm}{
\begin{tabular}{l|ccccccc}
\toprule
\multicolumn{1}{c|}{} &
  \multicolumn{3}{c|}{Image-to-Text} &
  \multicolumn{3}{c|}{Text-to-Image} &
   \\ \cline{2-7}
\multicolumn{1}{c|}{\multirow{-2}{*}{Method}} &
  {\color[HTML]{000000} R@1} &
  {\color[HTML]{000000} R@5} &
  \multicolumn{1}{c|}{{\color[HTML]{000000} R@10}} &
  {\color[HTML]{000000} R@1} &
  {\color[HTML]{000000} R@5} &
  \multicolumn{1}{c|}{{\color[HTML]{000000} R@10}} & \multirow{-2}{*}{rSum}  \\ \hline
 &
  \multicolumn{6}{c|}{1K Test} &
  \multicolumn{1}{l}{} \\
SCAN*\tiny{(18)\cite{lee2018stacked}} &
  72.7 &
  94.8 &
  \multicolumn{1}{c|}{98.4} &
  58.8 &
  88.4 &
  \multicolumn{1}{c|}{94.8} &
  507.9 \\
CVSE\tiny{(20)\cite{wang2020consensus}} &
  74.8 &
  95.1 &
  \multicolumn{1}{c|}{98.3} &
  59.9 &
  89.4 &
  \multicolumn{1}{c|}{95.2} &
  512.7 \\
IMRAM*\tiny{(20)\cite{chen2020imram}} &
  76.7 &
  95.6 &
  \multicolumn{1}{c|}{98.5} &
  61.7 &
  89.1 &
  \multicolumn{1}{c|}{95.0} &
  516.6 \\
SGRAF*\tiny{(21)\cite{diao2021similarity}} &
  79.6 &
  96.2 &
  \multicolumn{1}{c|}{98.5} &
  63.2 &
  90.7 &
  \multicolumn{1}{c|}{96.1} &
  524.3 \\
GPO\tiny{(21)\cite{chen2021learning}} &
  79.7 &
  96.4 &
  \multicolumn{1}{c|}{98.9} &
  64.8 &
  91.4 &
  \multicolumn{1}{c|}{96.3} &
  527.5 \\
  \begin{tabular}[c]{@{}l@{}}GPO+\\ SoftPool\tiny{(21)\cite{zeng2021softmax}}\end{tabular} &
  80.0 &
  96.5 &
  \multicolumn{1}{c|}{98.8} &
  65.2 &
  { 91.4} &
  \multicolumn{1}{c|}{{ 96.2}} &
  { 528.1} \\
CMCAN *\tiny{(22)\cite{zhang2022show}} &
  81.2 &
  \underline{96.8} &
  \multicolumn{1}{c|}{98.7} &
  65.4 &
  91.0 &
  \multicolumn{1}{c|}{96.2} &
  529.3 \\
VSRN++\tiny{(22)\cite{li2022image}} &
  77.9 &
  96.0 &
  \multicolumn{1}{c|}{98.5} &
  64.1 &
  91.0 &
  \multicolumn{1}{c|}{96.1} &
  523.6 \\
NAAF*\tiny{(22)\cite{zhang2022negative}} &
  80.5 &
  { 96.5} &
  \multicolumn{1}{c|}{{ 98.8}} &
  64.1 &
  90.7 &
  \multicolumn{1}{c|}{96.5} &
  527.1 \\
 \multicolumn{1}{l|}{MV-VSE\tiny{(22)\cite{li2022multi}}} &
  {80.4} &
  96.6 &
  \multicolumn{1}{c|}{\underline{99.0}} &
  { 64.9} &
  91.2 &
  \multicolumn{1}{c|}{96.0} &
  528.1 \\
GPO+CFM\tiny{(22)\cite{wei2022synthesizing}} &
  { 80.3} &
  96.7 &
  \multicolumn{1}{c|}{98.9} &
  { 65.2} &
  91.5 &
  \multicolumn{1}{c|}{96.4} &
  529.0 \\
  CMSEI*\tiny{(23)\cite{ge2023cross}} &
  { \underline{81.4}} &
  96.6 &
  \multicolumn{1}{c|}{98.8} &
  { \underline{65.8}} &
  \underline{91.8} &
  \multicolumn{1}{c|}{\textbf{96.8}} &
  \underline{531.2} \\
 \hline
\textbf{DCIN(ours)} &
  80.0 &
  96.7 &
  \multicolumn{1}{c|}{\textbf{99.0}} &
  65.2 &
  91.6 &
  \multicolumn{1}{c|}{96.2} &
  528.7 \\
\textbf{DCIN-ek(ours)} &
  80.9 &
  96.5 &
  \multicolumn{1}{c|}{98.8} &
  65.1 &
  91.5 &
  \multicolumn{1}{c|}{96.3} &
  529.1 \\
\textbf{DCIN*(ours)} &
  \textbf{81.4} &
  \textbf{96.8} &
  \multicolumn{1}{c|}{\textbf{99.0}} &
  \textbf{66.1} &
  \textbf{92.1} &
  \multicolumn{1}{c|}{\underline{96.6}} &
  \textbf{532.0} \\ \hline
 &
  \multicolumn{6}{c}{5K Test} &
  \multicolumn{1}{l}{} \\
SCAN*\tiny{(18)\cite{lee2018stacked}} &
  50.4 &
  82.2 &
  \multicolumn{1}{c|}{90.0} &
  38.6 &
  69.3 &
  \multicolumn{1}{c|}{80.4} &
  410.9 \\
IMRAM*\tiny{(20)\cite{chen2020imram}} &
  53.7 &
  83.2 &
  \multicolumn{1}{c|}{91.0} &
  39.7 &
  69.1 &
  \multicolumn{1}{c|}{79.8} &
  416.5 \\
GPO\tiny{(21)\cite{chen2020imram}} &
  58.3 &
  85.3 &
  \multicolumn{1}{c|}{92.3} &
  42.4 &
  72.7 &
  \multicolumn{1}{c|}{83.2} &
  434.3 \\
  \begin{tabular}[c]{@{}l@{}}GPO+\\ SoftPool\tiny{(21)\cite{zeng2021softmax}}\end{tabular} &
  59.4 &
  85.4 &
  \multicolumn{1}{c|}{\underline{92.8}} &
  42.6 &
  73.0 &
  \multicolumn{1}{c|}{83.3} &
  436.5 \\
VSRN++\tiny{(22)\cite{li2022image}} &
  54.7 &
  82.9 &
  \multicolumn{1}{c|}{90.9} &
  42.0 &
  72.2 &
  \multicolumn{1}{c|}{82.7} &
  425.4 \\
NAAF*\tiny{(22)\cite{zhang2022negative}} &
  58.9 &
  85.2 &
  \multicolumn{1}{c|}{92.0} &
  42.5 &
  70.9 &
  \multicolumn{1}{c|}{81.4} &
  430.9 \\
   \multicolumn{1}{l|}{MV-VSE\tiny{(22)\cite{li2022multi}}} &
  {59.1} &
  \underline{86.3} &
  \multicolumn{1}{c|}{92.5} &
  { 42.5} &
  72.8 &
  \multicolumn{1}{c|}{83.1} &
  436.3 \\
GPO+CFM\tiny{(22)\cite{wei2022synthesizing}} &
  59.7 &
  86.2 &
  \multicolumn{1}{c|}{92.2} &
  43.2 &
  73.2 &
  \multicolumn{1}{c|}{83.3} &
  437.8 \\
  CMSEI*\tiny{(23)\cite{ge2023cross}} &
  \textbf{61.5} &
  \underline{86.3} &
  \multicolumn{1}{c|}{92.7} &
  \underline{44.0} &
  \underline{73.4} &
  \multicolumn{1}{c|}{\underline{83.4}} &
  \underline{441.3} \\
 \hline
\textbf{DCIN(ours)} &
  59.1 &
  85.4 &
  \multicolumn{1}{c|}{92.2} &
  43.4 &
  73.7 &
  \multicolumn{1}{c|}{83.6} &
  437.4 \\
\textbf{DCIN-ek(ours)} &
  59.8 &
  85.8 &
  \multicolumn{1}{c|}{92.4} &
  42.9 &
  73.5 &
  \multicolumn{1}{c|}{83.6} &
  438.0 \\
\textbf{DCIN*(ours)} &
   \underline{60.8} &
  \textbf{86.3} &
  \multicolumn{1}{c|}{\textbf{93.0}} &
  \textbf{44.0} &
  \textbf{74.6} &
  \multicolumn{1}{c|}{\textbf{84.3}} &
  \textbf{443.0} \\
\bottomrule
\end{tabular}
}
\end{table}

\subsection{Comparison Results and Analysis}
We compare our DCIN model with the existing methods on two benchmark datasets. The results are shown in Table.~\ref{tab:1} and Table.~\ref{tab:2}, respectively. 
DCIN represents the use of causal inference alone, and DCIN-ek indicates the introduction of debiased external knowledge on DCIN. 
To ensure fairness in the experiments, we average two models’ similarities to report the ensemble results, denoted by DCIN*. The Table.~\ref{tab:1} shows the quantitative results on the Flickr30k dataset. Compared to CMSEI*~\cite{ge2023cross}, our DCIN* outperforms it in all evaluation metrics, achieving a 5.2\% increase in rSum. Among single models, DCIN-ek achieves the highest R@1 and obtains a 3.6\% improvement in rSum. Compared to the baseline GPO, DCIN-ek achieves 1.3\% and 1.9\% improvement in R@1 in two directions, respectively, and an 8\% increase in rSum. Moreover, the results on the larger and more complex MSCOCO dataset are presented in  Table.~\ref{tab:2}.
It can be observed that our method continues to outperform the existing methods in most evaluation metrics. Specifically, compared to the baseline GPO, our  DCIN* achieves improvements of 4.5\% and 8.7\% on the MSCOCO 1k test set and the full 5k test set, respectively. These improvements demonstrate the effectiveness of using causal inference to eliminate bias from the dataset and incorporate debiased external knowledge.

\begin{table}[]
\caption{The ablation study on the Flickr30K test set investigates the efficacy of each module in our approach.}
\label{tab:3}
\resizebox{\linewidth}{!}{
\setlength{\tabcolsep}{0.3mm}{
\begin{tabular}{l|ccc|ccc|c}
\toprule
\multicolumn{1}{c|}{} & \multicolumn{3}{c|}{Image-to-Text}            & \multicolumn{3}{c|}{Text-to-Image}   & {} \\ \cline{2-7}
\multicolumn{1}{c|}{\multirow{-2}{*}{Model}} & R@1  & R@5  & R@10 & R@1           & R@5  & R@10          & \multirow{-2}{*}{rSum}     \\ \hline
Base                  & 81.7 & 95.4 & 97.6 & 61.4          & 85.9 & 91.5          & 513.5 \\
\text{DCIN-ek}$^\dagger$               & 80.8 & 95.0 & 97.9 & \textbf{63.3} & 87.3 & 92.3          & 516.6 \\
DCIN                  & 82.2 & 96.0 & 98.0 & 62.9          & 87.5 & \textbf{92.5} & 519.1 \\
DCIN-ek                                       & \textbf{83.0} & \textbf{96.4} & \textbf{98.6} & \textbf{63.3} & \textbf{87.8} & 92.4 & \textbf{521.5}   \\ 
\bottomrule
\end{tabular}
}
}
\end{table}

\begin{table}[]
\caption{The ablation study on the Flickr30K test set investigates the generalizability of our method on different backbone architectures.}
\label{tab:5}
\resizebox{\linewidth}{!}{
\setlength{\tabcolsep}{0.3mm}{
\begin{tabular}{lccccccc}
\toprule
\multicolumn{1}{c|}{}                         & \multicolumn{3}{c|}{Image-to-Text}                                                                                    & \multicolumn{3}{c|}{Text-to-Image}                                                                                    &                                        \\ \cline{2-7}
\multicolumn{1}{c|}{\multirow{-2}{*}{Method}} & R@1                                   & R@5                                   & \multicolumn{1}{c|}{R@10}             & R@1                                   & R@5                                   & \multicolumn{1}{c|}{R@10}             & \multirow{-2}{*}{rSum}                 \\ \hline
VSRN\tiny{\cite{li2019visual}}                                          & 73.6                                  & 92.1                                  & 95.8                                  & 56.1                                  & 82.3                                  & 88.4                                  & 488.3                                  \\
VSRN+DCIN                                     & \textbf{74.6} & \textbf{92.8} & \textbf{96.1} & \textbf{58.5} & \textbf{84.4} & \textbf{90.4} & \textbf{496.8} \\ \hline
SVSE\tiny{\cite{zeng2021softmax}}                                           & 81.4                                  & 95.9                                  & 98.1                                  & 64.4                                  & 87.9                                  & 92.8                                  & 520.5                                  \\
SVSE+DCIN                                     &\textbf{82.5} & \textbf{96.2} & \textbf{98.6} & \textbf{64.9} & \textbf{88.2} & \textbf{93.6} & \textbf{524.0} \\ 
\bottomrule
\end{tabular}
}
}
\end{table}

\begin{table}[]
\caption{The ablation study on the Flickr30K test set investigates the performance of different sizes and modalities of dictionaries in our approach.}
\label{tab:4}
\resizebox{\linewidth}{!}{
\setlength{\tabcolsep}{0.3mm}{
\begin{tabular}{l|ccc|ccc|c}
\toprule
\multicolumn{1}{c|}{} & \multicolumn{3}{c|}{Image-to-Text} & \multicolumn{3}{c|}{Text-to-Image} &  \\ \cline{2-7}
\multicolumn{1}{c|}{\multirow{-2}{*}{Model}} & R@1  & R@5  & R@10 & R@1  & R@5  & R@10 & \multirow{-2}{*}{rSum}        \\ \hline
Base                  & 81.7 & 95.4 & 97.6 & 61.4 & 85.9 & 91.5 & 513.5 \\
DCIN w/o Init         & 80.1 & 95.5 & 97.8 & 62.7 & 86.7 & 92.2 & 515.0 \\
DCIN w/o VD\#300      & 82.1 & 95.1 & 98.0 & 62.8 & 86.9 & 92.0 & 516.9 \\
DCIN w/o LD\#300      & 81.4 & 95.4 & 98.1 & 62.3 & 87.5 & 92.5 & 517.2 \\
DCIN+D\#100           & 80.1 & 95.6 & 98.3 & 62.6 & 87.2 & \textbf{92.8} & 516.6 \\
DCIN+D\#200           & 80.9 & 96.0 & \textbf{98.7} & 62.9 & 87.3 & 92.5 & 518.3 \\
DCIN+D\#300       & 82.2 & 96.0 & 98.0 & 62.9 & 87.5 & 92.5 & 519.1 \\ 
DCIN+D\#400       & \textbf{82.4} & \textbf{96.1} & 98.4 & \textbf{63.2} & \textbf{87.7} & 92.2 & \textbf{520.0} \\ 
\bottomrule
\end{tabular}
}
}
\end{table}

\subsection{Ablation Study}\label{Ablation Study}
In this section, we perform several ablation studies on the Flickr30K dataset to evaluate the effectiveness of each module in our model, the influence of dictionary size as well as the different parameter settings. 

\subsubsection{The effectiveness of model component.}
In Table.~\ref{tab:3}, we examine the validity of each module. \textbf{Base:} We denote the GPO model as base model and compare it with the following variations. (1) \textbf{DCIN: }We incorporate causal inference into the base model. The experiments demonstrate that using the backdoor adjustment to eliminate bias in the dataset substantially enhances model performance, highlighting the effectiveness of confront visual and linguistic confounders to improve model generalization. (2) \textbf{DCIN-ek$^{\dagger}$: }We introduce biased external knowledge to reason    visual and linguistic dictionaries in our DCIN model. However, compared to DCIN, we observe a performance degradation, emphasizing that we cannot rely solely on relevance when incorporating external knowledge. (3) \textbf{DCIN-ek: }We introduced debiased external knowledge in DCIN, resulting in further performance improvements compared to DCIN. This indicates that introducing external knowledge can enhance model performance. However, considering (2), it is crucial to remove the bias of the dataset when introducing external knowledge to avoid negative effects on model generalization. 

Moreover, our method offers a plug-and-play capability that seamlessly integrates with existing approaches. When combined with VSRN \cite{li2019visual} and SVSE \cite{zeng2021softmax}, our DCIN further improves their performances (Table.~\ref{tab:5}), demonstrating its generalization ability. 

\subsubsection{The effectiveness of confounder dictionary.}
In Table. \ref{tab:4}, we perform exhaustive ablation studies to confirm the utility of the confounder dictionary. \textbf{DCIN w/o init: } We randomly initialize visual and linguistic confounder dictionaries with the size of 300. \textbf{DCIN w/o VD\#K: }We only use the linguistic confounder dictionary of size K. \textbf{DCIN w/o LD\#K: }We only use the visual confounder dictionary of size $K$. \textbf{DCIN+D\#K: }We define the size of the visual and linguistic confounder dictionaries as $K$. From the results we can find that even with random initialization of the dictionary, DCIN still outperforms the base model, indicating that random initialization of the dictionary can also reduce dataset bias \cite{yang2021causal} . Additionally, we find that a larger dictionary leads to better performance. Furthermore, we find that both visual and linguistic confounders introduce bias to the model, and removing either of them  from the confounder dictionary will result in a degradation of model performance.
\begin{figure}[htb]
\includegraphics[width=0.95\linewidth]{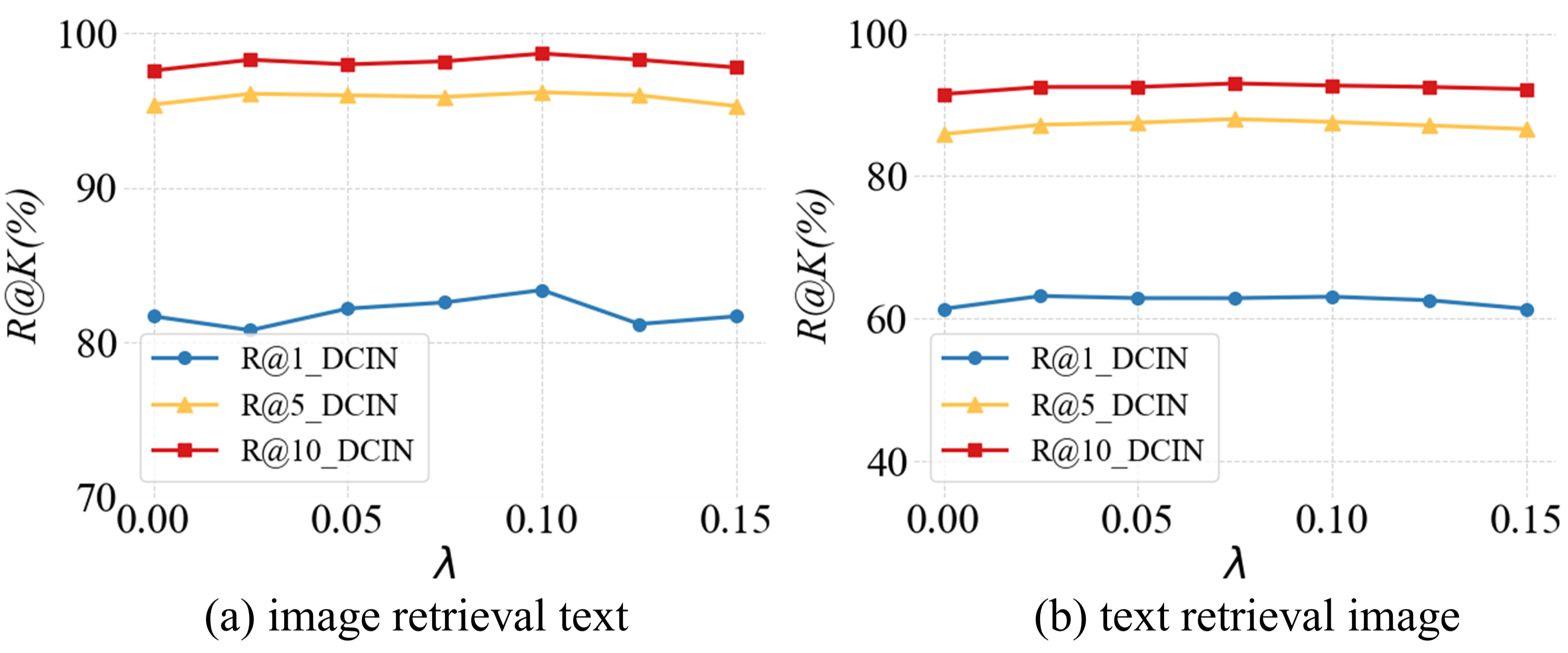}
\caption{The confidence level analysis for the approximate confounder dictionary on the Flickr30K test set. }
\label{fig:canshu}
\end{figure}

 \begin{figure}[htp]
  \includegraphics[width=0.90\linewidth]{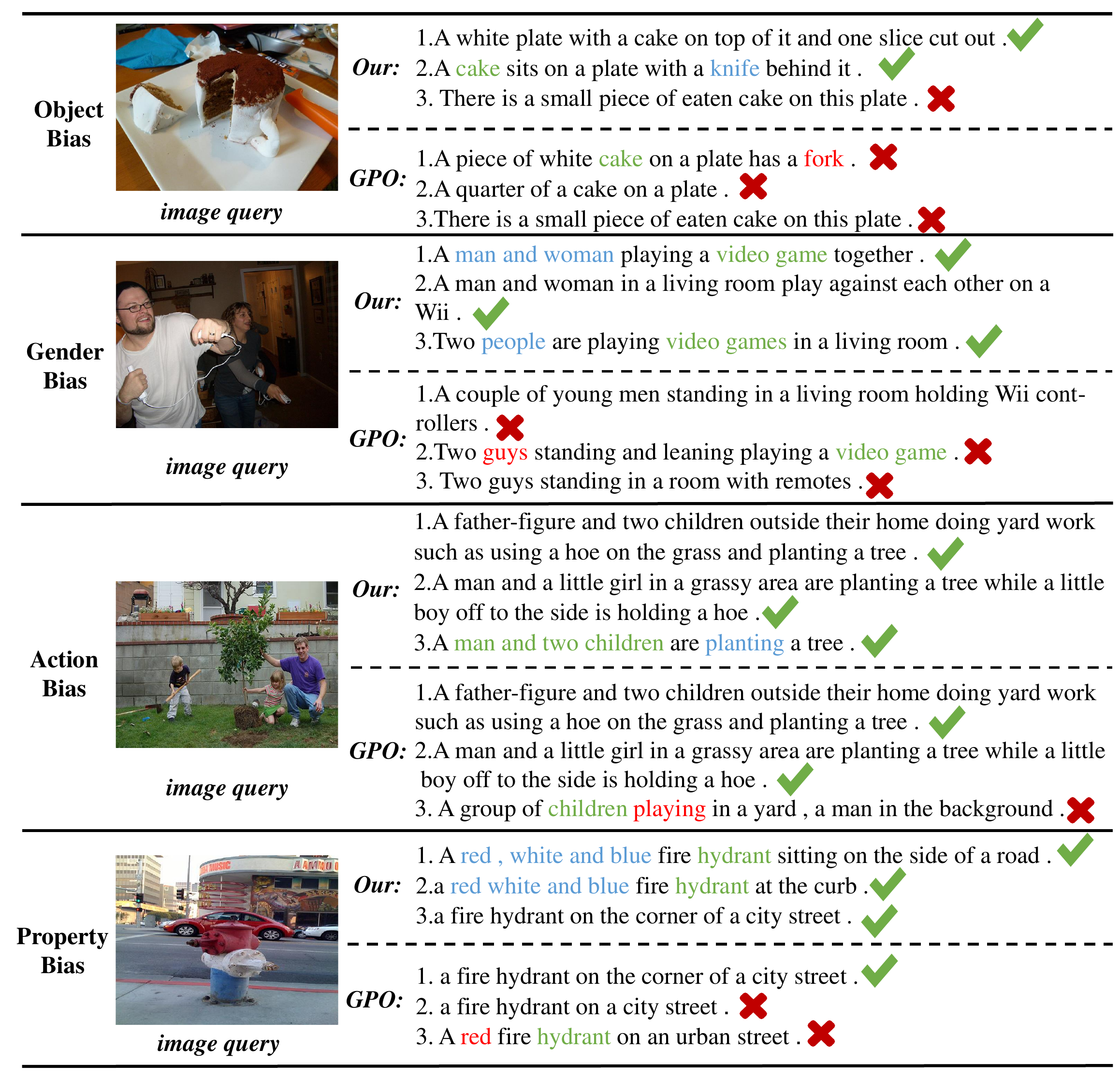}
  \caption{In the cases of object, gender, action, and property biases, our model and GPO retrieve several texts using images as query. The green context indicates concepts that may be biased by visual and linguistic confounders. The correct and incorrect text are marked with $\surd$  and $\times$, respectively.}
  \label{fig:case}
  \vspace{-0.1em}  
\end{figure}
\subsubsection{The analysis of confidence level $\lambda$.}
In Figure. \ref{fig:canshu}, we evaluate the impact of approximate sampling of visual and linguistic confounder dictionaries on image-text matching. We vary the confidence $\lambda$ from 0 to 0.15 with increments of 0.025 to balance the effect of approximation error on causal inference. The results show that the performance improves with increasing $\lambda$ until it reaches an optimal point. After that, there is a slight drop in performance, which demonstrates that this approximate visual and linguistic confounder dictionaries within suitable confidence levels can indeed  enhance the model's ability of inferring image-text causality.

\subsection{Qualitative results }
Figure.~\ref{fig:case} visualizes the image-to-text results retrieved by our model and the GPO. Intuitively, our model matches the texts that are less biased than the GPO. For instance, our model successfully reduces the spurious correlation between the "cake"  and the "fork", which is caused by inter-modal and intra-modal confounders. Moreover, our model also mitigates gender, action, and property biases, indicating that our DCIN can effectively eliminate visual and linguistic confounds. These results further confirm the effectiveness of our proposed approach.

\section{Conclusion}
In this paper, we provide an in-depth analysis of why image-text matching models are prone to learning spurious correlations from a causal perspective and propose a novel method called the Deconfounded Causal Inference Network for image-text matching. It utilizes layered interventions with two structural causal models in image-text matching and introduces external knowledge to mitigate spurious correlations. This strategy effectively addresses the issue of overestimating feature similarity between concepts and enhances the model's matching ability when introducing debiased external knowledge. It enables the model to learn causal knowledge from both the training and external databases simultaneously. Experimental results on two prominent datasets validate the superiority of our proposed method.

\begin{acks}
This work was supported in part by the  National Natural Science Foundation of China (U21B2024, 62202327) and the China Postdoctoral Science Foundation (2022M712369).
\end{acks}
\normalem
\bibliographystyle{ACM-Reference-Format}
\bibliography{sample-base}

\end{document}